\documentclass[10pt,twocolumn,letterpaper]{article}
\usepackage[dvipsnames, svgnames, x11names]{xcolor}
\usepackage{cvpr}              % To produce the CAMERA-READY version
\usepackage{graphicx}
\usepackage{amsmath}
\usepackage{amssymb}
\usepackage{booktabs}
\usepackage{array}
\usepackage{multirow}
\usepackage{blindtext}
\usepackage{caption}
\usepackage{pifont}
\usepackage{textcomp}
\usepackage{mathrsfs}
\usepackage{algorithm2e}
\usepackage{amsmath}
\usepackage{amssymb}
\usepackage{epstopdf}
\usepackage{wasysym}
\usepackage{balance}
\usepackage[normalem]{ulem}
\makeatletter

\usepackage{color}
\usepackage{calc}

\usepackage[pagebackref,breaklinks,colorlinks,citecolor=LimeGreen]{hyperref}

\usepackage[capitalize]{cleveref}
\crefname{section}{Sec.}{Secs.}
\Crefname{section}{Section}{Sections}
\Crefname{table}{Table}{Tables}
\crefname{table}{Tab.}{Tabs.}

\def\cvprPaperID{1213} % *** Enter the CVPR Paper ID here
\def\confName{CVPR}
\def\confYear{2022}

\title{Semantic-Aware Domain Generalized Segmentation}

\author{Duo Peng$^{1}$\quad Yinjie Lei$^{1,*}$\quad Munawar Hayat$^{2}$\quad Yulan Guo$^{3}$\quad Wen Li$^{4}$\\
$^{1}$Sichuan University\quad $^{2}$Monash University $^{3}$Sun Yat-sen University\quad\\ $^{4}$University of Electronic Science and Technology of China\\
{\tt\small duo\_peng@stu.scu.edu.cn}, {\tt\small yinjie@scu.edu.cn}, {\tt\small munawar.hayat@monash.edu}\\ 
{\tt\small guoyulan@sysu.edu.cn}, {\tt\small liwenbnu@gmail.com}
}

\begin{document}


%--------------------------figure 1
\twocolumn[{%
\renewcommand\twocolumn[1][]{#1}%
\maketitle
\begin{center}
    \centering
    \captionsetup{type=figure}
    \vspace{-4mm}
    \includegraphics[width=1\textwidth,height=7.2cm]{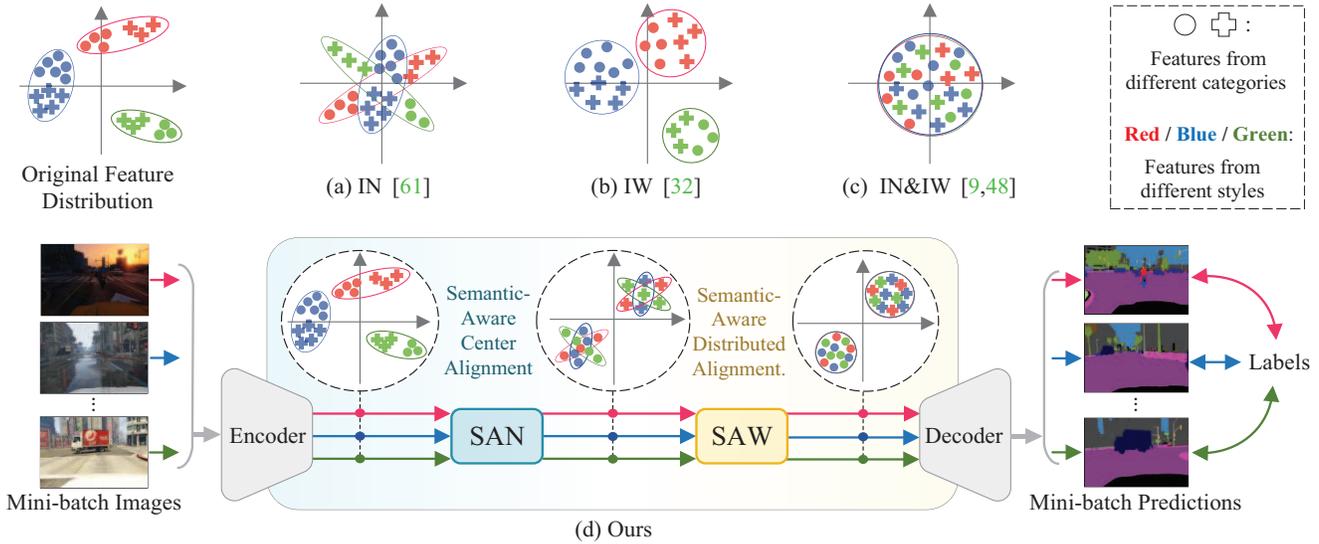}
    \captionof{figure}{Illustration of existing Instance Normalization and Whitening methods and our proposed approach. (a-c) Existing methods broadly eliminate the global distribution variance but ignore the category-level semantic consistency resulting in limited feature discrimination. (d) Our proposed modules (SAN \& SAW) encourage both intra-category compactness and inter-category separation through category-level feature alignment leading to both effective style elimination and powerful feature discrimination.}\label{fig1}
\end{center}%
}]
%--------------------------figure 1

%%%%%%%%% ABSTRACT
\begin{abstract}
Deep models trained on source domain lack generalization when evaluated on unseen target domains with different data distributions. The problem becomes even more pronounced when we have no access to target domain samples for adaptation. In this paper, we address domain generalized semantic segmentation, where a segmentation model is trained to be domain-invariant without using any target domain data. Existing approaches to tackle this problem standardize data into a unified distribution. We argue that while such a standardization promotes global normalization, the resulting features are not discriminative enough to get clear segmentation boundaries.\footnote{$^{\ast}$Corresponding Author: Yinjie Lei (yinjie@scu.edu.cn)} To enhance separation between categories while simultaneously promoting domain invariance, we propose a framework including two novel modules: 
\textbf{S}emantic-\textbf{A}ware \textbf{N}ormalization (\textbf{SAN}) and \textbf{S}emantic-\textbf{A}ware \textbf{W}hitening (\textbf{SAW}). 
Specifically, SAN focuses on category-level center alignment between features from different image styles, while SAW enforces distributed alignment for the already center-aligned features. With the help of SAN and SAW, we encourage both intra-category compactness and inter-category separability. We validate our approach through extensive experiments on widely-used datasets (\ie GTAV, SYNTHIA, Cityscapes, Mapillary and BDDS). Our approach shows significant improvements over existing state-of-the-art on various backbone networks. Code is available at \href{https://github.com/leolyj/SAN-SAW}{https://github.com/leolyj/SAN-SAW}

% The demo is uploaded in supplementary material and the code will be released upon the acceptance of this paper. %I suggest  you just write, "Our Codes will be released upon publication." dont confuse reviewers with your current statement--you're inviting an un-necassary doubt.. why demo now,etc   

\end{abstract}

%%%%%%%%% BODY TEXT
\section{Introduction}
\label{sec:intro}

Semantic segmentation is a critical machine vision task with multiple downstream applications, such as robotic navigation \cite{kim2018indoor,miyamoto2019vision,ye20183d,liu2020semantic}, autonomous vehicles \cite{geiger2012we,kumar2021syndistnet,yang2018denseaspp,peng2021sparse} and scene parsing \cite{zhang2020rapnet,zhang2020deep,zhang2017scale,lei2020hierarchical}. While the current fully supervised deep learning based segmentation methods can achieve promising results when they are trained and evaluated on data from same domains \cite{long2015fully,chen2014semantic,chen2017rethinking,chen2017deeplab,he2017mask,badrinarayanan2017segnet,zhang2019deep,ma2020global,zhang2020semantic}, their performance dramatically degrades when they are evaluated on unseen out-of-domain data. 
To enable generalization of models across domains, different domain adaptation techniques have been recently proposed \cite{ben2007analysis,ganin2015unsupervised,ganin2016domain,hoffman2018cycada,murez2018image,pan2020unsupervised,saito2018maximum,vu2019advent,zou2018unsupervised}. However, a critical limitation of domain adaptation methods is their reliance on the availability of target domain in advance for training purposes. This is impractical for many real-world applications, where it is hard to acquire data for rarely occurring concepts.

In this paper, we consider the challenging case of \textit{\textbf{D}omain \textbf{G}eneralized \textbf{S}emantic \textbf{S}egmentation (\textbf{DGSS})}, where we do not have access to any target domain data at the training time \cite{choi2021robustnet,pan2018two,pan2019switchable,peng2021global,yue2019domain}. %This introduces multiple challenges such as overfitting, training aimlessness, and inapplicability of traditional knowledge transfer techniques.
Existing methods tackle DGSS using two main approaches: (1) \textit{Domain Randomization} \cite{yue2019domain,peng2021global} which aims to increase the variety of training data by augmenting the source images to multiple domain styles. However, this is limiting since the augmentation schemes used are unable to cover different scenarios that may occur in the target domain. (2) \textit{Normalization and Whitening} \cite{pan2018two,pan2019switchable,choi2021robustnet} which utilizes predefined Instance Normalization (IN) \cite{ulyanov2017improved} or Instance Whitening (IW) \cite{li2017universal} to standardize the feature distribution of different samples.
%In order to model a variety of commonly observed appearance variations within images (see Fig. \ref{fig2}),
IN separately standardizes features across each channel of individual images to alleviate the feature mismatch caused by style variations. However, as shown in~\cref{fig1} (a), IN only achieves center-level alignment and ignores the joint distribution among different channels. IW can remove linear correlation between channels, leading to well-clustered features of uniform distributions (see~\cref{fig1} (b)). Recent studies \cite{pan2019switchable,choi2021robustnet} propose to combine IN and IW to achieve joint distributed feature alignment (see~\cref{fig1} (c)). Nevertheless, such global alignment strategy lacks the consideration of local feature distribution consistency. The features belonging to different object categories, which are originally well separated, are mapped together after normalization, leading to confusion among categories especially when generalizing to unseen target domains. Such semantic inconsistency inevitably results in sub-optimal training, causing performance degradation on unseen target domain and even the training domain (\ie source domain).

To address the inherent limitations of IN and IW, we propose two modules, Semantic-Aware Normalization (SAN) and Semantic-Aware Whitening (SAW), which collaboratively align category-level distributions aiming to enhance the discriminative strength of features (see~\cref{fig1} (d)).
Compared with traditional IN\&IW based methods, our approach brings two appealing benefits:
\textit{First}, it carefully integrates semantic-aware center alignment and distributed alignment, enabling both discriminative and compact matching of features from different styles. Therefore, our method can significantly enhance models' generalization to out-of-domain distributed data.
\textit{Second}, existing methods improve the generalization ability at the cost of source domain performance \cite{pan2018two,choi2021robustnet}. Nevertheless, our approach enhances the semantic consistency while improving category-level discrimination, thus leading to effective generalization with negligible performance drop on source domain.

% Therefore, our method can discriminatively and compactly map features belonging to different styles, thus significantly enhancing models generalization to out-of-domain distributed data.

% Second, it provides a self-adaptive local feature matching which naturally achieves global distribution alignment, thus enhancing model's generalization capability. %Our empirical evaluations show that our method significantly boost the DGSS performance with negligible drop in test set of source domain.

Our extensive empirical evaluations on benchmark datasets show that our approach improves upon previous DGSS methods, setting new state-of-the-art performances. Remarkably, our method also performs favorably compared with existing SOTA domain adaptation methods that are trained using target domain data. In summary, followings are the major contributions of our work.

\vspace{-1.5mm}
\begin{itemize}
\setlength{\itemsep}{1pt}
\setlength{\parsep}{1pt}
\setlength{\parskip}{1pt}
%\item \pd{We propose an effective approach to alleviate the semantic mismatch problem of existing global alignment methods by developing two novel feature alignment modules for semantic-aware feature centering and unification.}
\item We propose effective feature alignment strategies to tackle out-of-domain generalization for segmentation, without access to target domain data for training.
% We propose an effective approach to tackle out-of-domain generalization for semantic segmentation, considering the challenging scenario where the target domain data is not available for training. Our proposed approach alleviates the limitations of existing methods by developing two novel feature alignment modules which sequentially centre and unify category-level features, to promote generalization.
\item The proposed semantic-aware alignment modules, SAN and SAW, are plug-and-play and can easily be integrated with different backbone architectures, consistently improving their generalization performance.
\item Through extensive empirical evaluations, and careful ablation analysis, we show the efficacy of our approach across different domains and backbones, where it significantly outperforms the current state-of-the-art. Remarkably, we even perform at par with approaches using target domain data for training purposes.

% Our approach achieves new state-of-the-art performance on benchmark datasets.

\end{itemize}

% %--------------------------figure 2
%     \begin{figure}[t]
%     \centering{}\vspace{-0mm}
%      \includegraphics[scale=0.4]{figtab/appear diverse.pdf} 
%      \caption{Images show different appearance styles even within a same dataset. Style variance is pervasive in datasets since the change of illumination and scenarios caused by the long time span of image capture. We show samples from GTAV, BDDS, Mapillary and SYNTHIA which are widely used in DGSS.
%      }
%     \label{fig2}\vspace{-2mm}
    
%     \end{figure}
% %--------------------------figure 2

%-------------------------------------------------------------------------
\section{Background}

Here, we discuss recent approaches developed for Domain Adaptation (DA) and Domain Generalization (DG) in the context of semantic segmentation. %Therefore, we mainly review the DA and DG approaches for semantic segmentation here.

\subsection{Domain Adaptation (DA)}

Domain Adaptation seeks to narrow the domain gap between the source and target domain data. It aims to enhance the generalization ability of the model by aligning the feature distributions between the source and target images \cite{ganin2015unsupervised,ganin2016domain,long2015learning,long2016unsupervised,sohn2017unsupervised,tzeng2015simultaneous,tzeng2017adversarial}. Domain adaptation for semantic segmenation (DASS) was first studied in  \cite{hoffman2016fcns,zhang2017curriculum}, and since then has gained significant research attention. We can broadly categorize the existing approaches for DASS into Adversarial Training, and Self-Training based methods.
Most of the existing work on DASS has been dominated by \textit{Adversarial Training} based approaches \cite{ganin2015unsupervised,hoffman2018cycada,chen2018road,sankaranarayanan2018learning,zhang2018fully}. Inspired by Generative Adversarial Networks \cite{goodfellow2020generative}, these approaches are generative in nature, and synthesize indistinguishable features which are domain-invariant and deceive the domain classifier.
\textit{Self-Training} based DASS approaches are relevant once labeled training data is scarce. These methods \cite{zou2018unsupervised, li2019bidirectional} train the model with pseudo-labels which are generated from the previous models predictions. 
However, DA methods require access to the samples from the target domain, which limits their applicability on totally unseen target domain.

\subsection{Domain Generalization (DG)}

In contrast to Domain Adaptation, where the images in the target domain, although without labels, are accessible during the training process, Domain Generalization is evaluated on data from totally unseen domains \cite{muandet2013domain,gan2016learning}. Domain generalization has been mostly explored on the image classification task, and a number of approaches have been proposed using as meta-learning \cite{li2018learning,balaji2018metareg,li2019episodic,li2019feature}, adversarial training \cite{li2018domain,li2018deep,rahman2020correlation}, autoencoders \cite{ghifary2015domain,li2018deep}, metric learning \cite{dou2019domain,motiian2017unified} and data augmentation \cite{gong2019dlow,zhou2020learning}. 
The research on domain generalization for semantic segmentation (DGSS) is still in its infancy, with only a few existing approaches \cite{yue2019domain,peng2021global,pan2018two,pan2019switchable,choi2021robustnet}. These existing DGSS methods mainly focus on two aspects: (1) Domain Randomization and (2) Normalization and Whitening. \textit{Domain Randomization} based methods seek to synthesize images with different styles \eg \cite{yue2019domain} leverages the advanced image-to-image translation to transfer a source domain image to multiple
styles aiming to learn a model with high generalizability. Similarly, GTR \cite{peng2021global} randomizes the synthetic images with the styles of unreal paintings in order to learn domain-invariant representations.
\textit{Normalization and Whitening Methods} apply different normalization techniques such as Instance Normalization (IN) \cite{ulyanov2017improved} or whitening \cite{pan2019switchable}. For example, based on the observation that Instance Normalization (IN) \cite{ulyanov2017improved} prevents overfitting on domain-specific style of training data, \cite{pan2018two} proposes to utilize IN to capture style-invariant information from appearance changes while preserving content related information. Inspired from \cite{pan2018two}, \cite{pan2018two} proposes Switchable Whitening (SW), which combines IN with other whitening methods, aiming to achieve a flexible and generic features. In another recent approach \cite{choi2021robustnet}, an instance selective whitening to disentangle domain-specific and domain-invariant properties is explored and only domain-specific features are normalized and whitened. However, all aforementioned methods perform a global alignment for features belong to different image categories. We aim to address this crucial limitation and propose an approach which enforces local semantic consistency during the trend of global style elimination.

\section{Preliminaries}

% We focus on the problem of category feature alignment in Domain Genralized Semantic Segmentation (DGSS), where we only have access to the source data $X_{\mathrm{S}}$ with pixel-level labels $Y_{\mathrm{S}}$. Our goal is to adopt feature constraint on middle-layer features $\mathbf{F}$ when training on source labeled data, aiming to train a model $G$ that can reliably extract domain-invariant features. 

Let's denote an intermediate mini-batch feature map by $\mathbf{F}\in\mathbb{R}^{N\times K\times H\times W}$, where $N$, $K$, $H$ and $W$ are the dimensions of the feature map, \ie \textit{batch sample}, \textit{channel}, \textit{height} and \textit{width}, respectively.
$\mathbf{F}_{n,k,h,w} \in \mathbf{F}$ represent the feature element, where $n$, $k$, $h$, $w$ respectively indicate the index of each dimension.
Similarly, $\mathbf{F}_{n}\in\mathbb{R}^{K\times H\times W}$ denotes the features of $n$-th sample from mini-batch, and $\mathbf{F}_{n,k}\in\mathbb{R}^{H\times W}$ denotes the $k$-th channel of $n$-th sample. 

Below, we first define Instance Normalization (IN) and Instance Whitening (IW), which have been commonly used by the existing approaches.

% Given an intermediate mini-batch feature map $\mathbf{F}\in\mathbb{R}^{N\times K\times H\times W}$, let $\mathbf{F}_{n,k,h,w}$ denote the feature element, where $n$, $k$, $h$, $w$ indicate the index of \textit{sample}, \textit{channel}, \textit{height}, and \textit{width} respectively. Similarly, $\mathbf{F}_{n}\in\mathbb{R}^{K\times H\times W}$ denotes the features of $n$-th sample from mini-batch, and $\mathbf{F}_{n,k}\in\mathbb{R}^{H\times W}$ denotes the $k$-th channel of $n$-th sample. 

% Traditional normalization methods consider two aspects to handle DGSS, Instance Normalization (IN) and Instance Whitening (IW), which standardize features along \textit{\textbf{spatial}} and \textit{\textbf{channel}} respectively. 

% Next, we give detailed analysis on the problem of each aspect, in order to show the motivation of our feature constraint on spatial and channel axes.

% Let $\mathbf{F}\in\mathbb{R^{\mathit{N\times KHW}}}$ be the feature matrix of a mini-batch, where $N$, $K$, $H$, $W$ indicate the number of samples, number of channels, height, and width respectively. Thus, we use matrix $\mathbf{F}_{n}\in\mathbb{R^{\mathit{K\times HW}}}$ to represent the $n$-th sample of the mini-batch, where $n\in\{1,2,...,N\}$. Our goal is to adopt feature constraint on middle-layer features $\mathbf{F}$ when training on source labeled data, aiming to train a model $G$ that can reliably extract domain-invariant features.

\textbf{Instance Normalization (IN)} simply standardizes features using statistics (\ie mean and standard deviation) computed over each individual channel from each individual sample, given by:
\begin{equation}
\mathrm{IN}(\mathbf{F})=\frac{\mathbf{F}_{n,k}-\mu_{n,k}}{\sigma_{n,k}+\varepsilon},
\end{equation}
where $\mathrm{IN}(\cdot)$ denotes the instance normalization process and $\varepsilon$ is a small value to avoid division by zero. The mean $\mu_{n,k}$ and standard deviation $\sigma_{n,k}$ of $n$-th sample $k$-th channel are computed as follows:
\vspace{-2mm}
\begin{equation}
\mu_{n,k}=\frac{1}{HW}\sum_{h=1}^{H}\sum_{w=1}^{W}\mathbf{F}_{n,k,h,w},
\end{equation}
\begin{equation}
\sigma_{n,k}=\sqrt{\frac{1}{HW}\sum_{h=1}^{H}\sum_{w=1}^{W}(\mathbf{F}_{n,k,h,w}-\mu_{n,k})^{2}}.
\end{equation}

Using above operations, IN transforms the features from different image samples to have a standard distribution, i.e, zero mean and one standard deviation. However, even though the features of each channel are centered and scaled into standard distribution, the joint distribution between channels might be mismatched.
% To this end, we consider to distinguish the features of different categories in the spatial dimension, so as to standardize each category features respectively, thus achieving semantic consistency for better DGSS performance.

%--------------------------figure 2
    \begin{figure*}[t]
    \centering{}\vspace{-0mm}
     \includegraphics[scale=0.58]{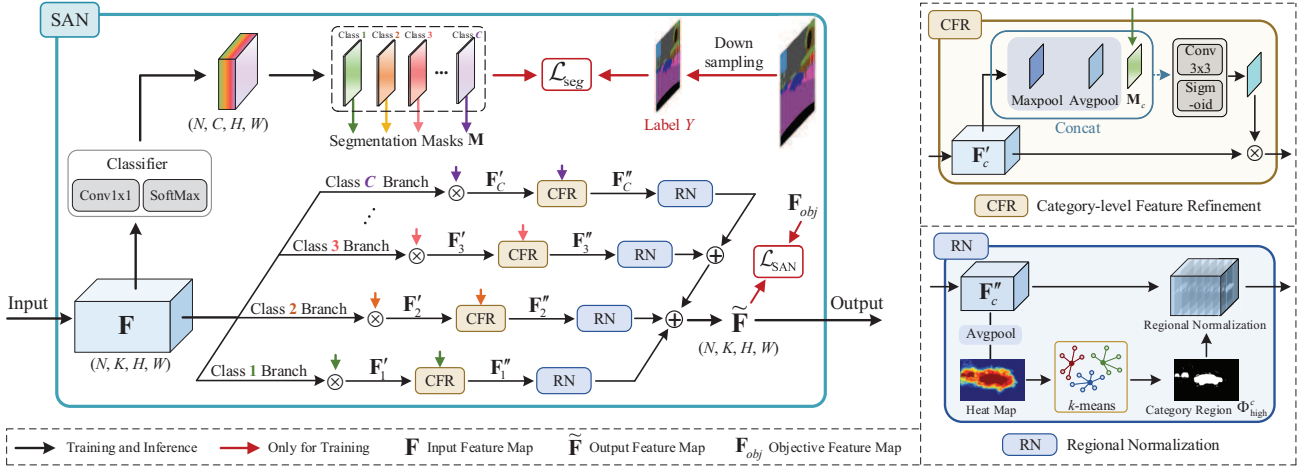} 
     \caption{The detailed architecture of our Semantic Aware Normalization (SAN) module. SAN adapts a multi-branch normalization strategy, aiming to transform the feature map $\mathbf{F}$ into the category-level normalized features $\widetilde{\mathbf{F}}$, that are  semantic-aware center aligned.
     }
    \label{fig3}\vspace{-4mm}
    
    \end{figure*}
%--------------------------figure 2

%-------------------------------------------------------------------------

% Numerous studies [13, 14, 30, 56, 51, 36, 7, 45, 57] show that whitening features of individual image sample is able to filter out information of image appearance because channel correlation contains domain-specific style such as texture and color. With the whitening on each sample, the style difference between samples is eliminated. 

\textbf{Instance Whitening (IW)} standardizes features by decorrelating the channels. As shown in \cref{fig4} (a), it removes correlation between channels by making the covariance matrix close to the identity matrix through the following objective function:
\vspace{-3mm}
\begin{equation}
\mathcal{L}_{\mathrm{IW}}=\sum_{n=1}^{N}||\Psi(\mathbf{F}_{n})-\mathbf{I}||_{1},\label{eq:4}
\end{equation}
where $\Psi(\cdot)$ and $\mathbf{I}$ denotes the channel correlation and identity matrix. $\Psi(\mathbf{F}_{n})$ is defined as:
\vspace{1mm}
\begin{footnotesize} 
\begin{equation}
\Psi(\mathbf{F}_{n})=\begin{bmatrix}\mathrm{Cov}(\mathbf{F}_{n,1},\mathbf{F}_{n,1}) & \cdots & \mathrm{Cov}(\mathbf{F}_{n,1},\mathbf{F}_{n,K})\\
\vdots & \ddots & \vdots\\
\mathrm{Cov}(\mathbf{F}_{n,K},\mathbf{F}_{n,1}) & \cdots & \mathrm{Cov}(\mathbf{F}_{n,K},\mathbf{F}_{n,K})
\end{bmatrix},
\end{equation}
\end{footnotesize}
\vspace{1mm}
where $\mathrm{Cov}(\mathbf{F}_{n,i},\mathbf{F}_{n,j})$ is the covariance value between the $i$-th channel and $j$-th channel of feature $\mathbf{F}_{n}$, given by:
\begin{footnotesize}
\begin{equation}
\mathrm{Cov}(\mathbf{F}_{n,i},\mathbf{F}_{n,j})=\frac{1}{HW}\sum_{h=1}^{H}\sum_{w=1}^{W}(\mathbf{F}_{n,i,h,w}-\mu_{n,i})(\mathbf{F}_{n,j,h,w}-\mu_{n,j}).
\end{equation}
\end{footnotesize} 

IW \cite{li2017universal} is capable of unifying the joint distribution shape through channel decorrelation for each sample.
Combined with IN \cite{ulyanov2017improved}, IW can make a unified joint distributed alignment. However, such global matching might cause some features to be mapped to an incorrect semantic category, resulting in poor segmentation boundary decisions. In the following \cref{sec:proposed method}, we introduce our method which aims to tackle these problems, and preserve the semantic relationships between different categories.

% Moreover, such strong constraint which strictly eliminates correlation between all channels may damage the semantic content resulting in the loss of important domain-invariant information. To address above issues, we consider to find the most related channels for each category aiming to carry out category-level whitening which is essentially moderate local channel constraint for DGSS.

% It is well-known that channel features are also highly related to categories. When making segmentation decision for a specific category, only several channels play an important role. 

% However, Instance Whitening globally normalizes all channels without distinction, facing the same problem of category semantic inconsistency.

%-------------------------------------------------------------------------
\section{Proposed Method \label{sec:proposed method}}

% This section presents our approach to solve semantic inconsistency problem in DGSS. 
Our goal is to achieve \textbf{semantic-aware center alignment} and \textbf{distributed alignment}. For this, we introduce two novel modules, \textit{\textbf{S}emantic-\textbf{A}ware \textbf{N}ormalization (SAN)} and \textit{\textbf{S}emantic-\textbf{A}ware \textbf{W}hitening (SAW)}. We sequentially embed these two modules in our network as shown in \cref{fig1}. We discuss these modules in detail below.%^ the proposed SAN and SAW modules in detail.

% Since DNNs embed informative cues into both spatial and channel features, we adopt feature constraint modules to extract meaningful domain-invariant features along those two principal dimensions: channel and spatial axes. To achieve this, we sequentially apply spatial and channel feature distributed constraint (as shown in Fig. \ref{fig1}), so as to match the distributions precisely, thus increasing the domain generalization ability. The following describes the details of each category-level feature constraint module.

\subsection{Semantic-Aware Normalization (SAN) \label{subsec:Spatial Feature Constraint (SFC)}}
% We propose the Spatial Feature Constraint module for category-level distributed alignment along spatial axis.
Given an intermediate mini-batch feature map $\mathbf{F}$, SAN transforms $\mathbf{F}$ into a feature map that is category-level centred. With the help of segmentation labels $Y$, we can easily obtain the desired objective feature map $\mathbf{F}_{obj}$ as:
\vspace{-2mm}
\begin{equation}
\mathbf{F}_{obj}=\frac{\mathbf{F}_{n,k}^{c}-\mu_{n,k}^{c}}{\sigma_{n,k}^{c}+\varepsilon}\cdot\gamma^{c}+\beta^{c},\label{eq:7}
\end{equation}
\begin{equation}
\mu_{n,k}^{c}=\frac{1}{|Y(c)|}\sum_{Y(c)}\mathbf{F}_{n,k}^{c},
\end{equation}
\begin{equation}
\sigma_{n,k}^{c}=\sqrt{\frac{1}{|Y(c)|}\sum_{Y(c)}(\mathbf{F}_{n,k}^{c}-\mu_{n,k}^{c})^{2}},
\end{equation}
\noindent where $\mu_{n,k}^{c}$ and $\sigma_{n,k}^{c}$ are the mean and standard deviation computed from $c$-th category features of $k$-th channel, $n$-th sample, and the $c$-th category label $Y(c)$. Features $\mathbf{F}_{n,k}^{c}$ belong to $c$-th category in channel $\mathbf{F}_{nk}$. The weights for scaling and shifting are denoted by $\gamma$ and $\beta$, respectively, which are both learnable parameters. We seperately allocate these affine parameters for each category (\ie $\gamma^{c}$ and $\beta^{c}$) aiming to adjust the standardized features from different categories to distinct spaces, thus making our feature space more discriminative. Note that different samples in the mini-batch share the same affine parameters in order to cast features of same category into a same feature space, thus ensuring category-level center alignment.

% Equation (7) is essentially a category normalization which normalizes features for each spatial category region in each individual channel. 

%--------------------------figure 3
    \begin{figure*}[t]
    \centering{}\vspace{-2mm}
     \includegraphics[scale=0.49]{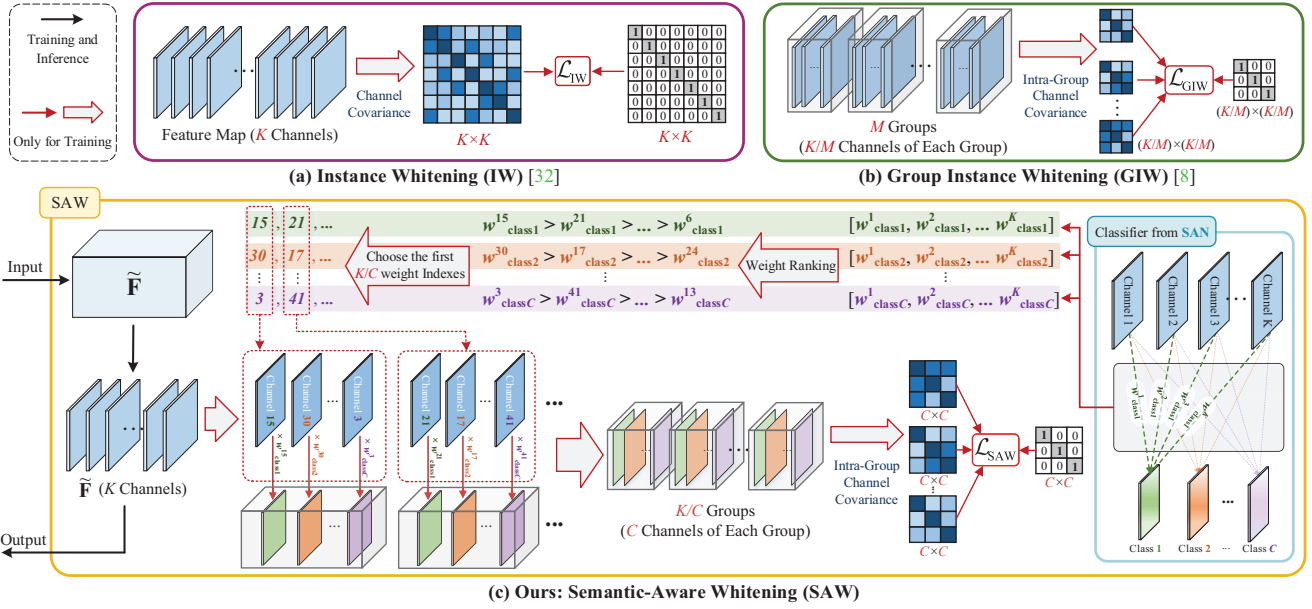} 
     \caption{Illustration of feature whitening in IW \cite{li2017universal}, GIW \cite{cho2019image} and the proposed SAW. \textbf{(a)} IW de-correlates all channels from each other. \textbf{(b)} GIW only de-correlates the channels in the same group. \textbf{(c)} SAW allocates channels related to different categories in each group.
     }
    \label{fig4}\vspace{-4mm}
    
    \end{figure*}
%--------------------------figure 3

We utilize SAN to approximate \cref{eq:7} since the label $Y$ is unavailable when testing on target domains. We follow a series of steps to transform the input features $\mathbf{F}$ into the objective features $\mathbf{F}_{obj}$ as shown in \cref{fig3}. First, we leverage the segmentation masks generated from the classifier to highlight the category region while simultaneously suppressing other regions in each normalization branch. 
\vspace{-0.8mm}
\begin{equation}
\mathbf{F}_{c}^{\prime}=\mathbf{F}\otimes\mathbf{M}_{c},
\end{equation}
where $\mathbf{M}_{c}$ denotes the mask of the $c$-th category, $\otimes$ denotes Hadamard product and $\mathbf{F}_{c}^{\prime}$ represents the masked feature map in $c$-th category branch. The generated mask can be too rough to precisely locate category features. We therefore introduce a Category-level Feature Refinement (CFR) block to adaptively adjust the highlighted features $\mathbf{F}_{c}^{\prime}$ into $\mathbf{F}_{c}^{\prime\prime}$, given by:
\begin{equation}
\mathbf{F}_{c}^{\prime\prime}=\mathrm{Sigm}(f^{3\times3}([\mathbf{F}_{c,max}^{\prime};\mathbf{F}_{c,avg}^{\prime};\mathbf{M}_{c}]))\otimes\mathbf{F}_{c}^{\prime},
\end{equation}
where $f^{3\times3}(\cdot)$ and $\mathrm{Sigm}(\cdot)$ denote $3$$\times$$3$ convolution and sigmoid function, respectively. $\mathbf{F}_{c,max}^{\prime}$  and $\mathbf{F}_{c,avg}^{\prime}$ are max-pooled and average-pooled features of feature map $\mathbf{F}_{c}^{\prime}$. $[\mathbf{a};\mathbf{b}]$ is the concatenation of $\mathbf{a}$ and $\mathbf{b}$ along channel axis.

In order to further refine the category-level center alignment, we design a Regional Normalization layer which normalizes features only within the category region instead of whole scene. After refinement, only the feature elements with high value are assigned the category region. To flexibly identify feature elements, we apply $k$-means clustering on the spatial feature map obtained by avgpooling along channel axis. After dividing the spatial elements into $k$ clusters, the clusters from the first to the $t$-th are considered to be the category region, and the remaining clusters are considered as ignored region. We set $t$ to one and search the optimal $k$ through the hyper-parameter search. In this paper, $k$ is set to 5. See \cref{sec: hyper-param} for more details.

Thus, we can assign the feature elements of $c$-th branch into 2 clusters $\{\mathrm{\Phi}_{\mathrm{low}}^{c},\mathrm{\Phi}_{\mathrm{high}}^{c}\}$, where $\mathrm{\Phi}_{\mathrm{high}}^{c}$ denotes the identified category region. We normalize features within the category region $\mathrm{\Phi}_{\mathrm{high}}^{c}$ for each individual channel. Finally, all category branches are added together, then re-shifted and scaled by the learnable affine parameters:
\vspace{-0.8mm}
\begin{equation}
\mathbf{\widetilde{F}}=\sum_{c=1}^{C}\mathrm{RN}(\mathbf{F}_{c}^{\prime\prime},\Phi_{\mathrm{high}}^{c})\cdot\gamma^{c}+\beta^{c},
\end{equation}
where $\mathrm{RN}(\cdot,\Phi_{\mathrm{high}}^{c})$ denotes our regional normalization in $c$-th branch, $\gamma^{c}$ and $\beta^{c}$ are affine parameters as per \cref{eq:7}.
In order to ensure the processed feature map $\widetilde{\mathbf{F}}$ are category-level-center-aligned as per \cref{eq:7}, we optimize the following cross-entropy loss:
\vspace{-0.8mm}
\begin{equation}
\mathcal{L}_{\mathrm{SAN}}=\mathrm{CE}(\mathbf{M},Y)+||\mathbf{\widetilde{F}}-\mathbf{F}_{obj}||_{1},
\end{equation}
where $\mathbf{M}\in\{\mathbf{M}_{1},\mathbf{M}_{2},...,\mathbf{M}_{C}\}$ denotes the set of predicted segmentation masks.% and $\mathrm{CE}(\cdot,\cdot)$ denotes the cross entropy loss.

\subsection{Semantic-Aware Whitening (SAW) \label{subsec:Channel Feature Constraint (CFC)}}

%  However, such strong eliminates all channels' correlation with each other leading to limited DGSS performance.

% As shown in Fig.~\ref{fig1}, the 

% As shown in Fig.~\ref{fig1}, the Semantic-Aware Whitening (SAW) module is proposed to further enhance channel decorrelation for the distributed alignment of the already class-wise centred features. Instance Whitening (IW) is capable to unify the joint distribution, which is useful for distributed alignment. However, directly adopting IW is not feasible, since such strong whitening, which strictly removes correlation between all channels, may damage the semantic content, resulting in the elimination of important domain-invariant information. We find a straightforward way to solve the above problem: Group Instance Whitening (GIW, shown in Fig. \ref{fig4} (b)). Given the feature map $\widetilde{\mathbf{F}}$ which is the output of SAN module, GIW effectively improve the generalization performance by partial (group) channel decorrelation:

We propose the Semantic-Aware Whitening (SAW) module, to further enhance channel decorrelation for the distributed alignment of the already semantic-centred features.
Instance Whitening (IW) is capable of unifying the joint distribution, which is useful for distributed alignment. However, directly adopting IW is not feasible, since such strong whitening that strictly removes correlation between all channels may damage the semantic content, resulting in loss of crucial domain-invariant information. Group Instance Whitening (GIW \cite{cho2019image}, shown in \cref{fig4} (b)) is a simple solution to this problem. Given the feature map $\widetilde{\mathbf{F}}$ which is the output of SAN module, GIW is defined by:
\vspace{-0.8mm}
\begin{equation}
\mathbf{G}_{n}^{m}=[\mathbf{\widetilde{F}}_{n,\frac{K(m-1)}{M}+1};\thinspace\mathbf{\widetilde{F}}_{n,\frac{K(m-1)}{M}+2};\thinspace...\thinspace;\thinspace\mathbf{\widetilde{F}}_{n,\frac{Km}{M}}],\label{eq:14}
\end{equation}
\begin{equation}
\mathcal{L}_{\mathrm{GIW}}=\frac{1}{N}\sum_{n=1}^{N}\sum_{m=1}^{M}||\Psi(\mathbf{G}_{n}^{m})-\mathbf{I}||_{1},
\end{equation}
where $\mathbf{G}_{n}^{m}$ denotes the $m$-th group of $n$-th sample, $M$ is the number of groups, and $\Psi(\cdot)$ is the channel correlation matrix defined in \cref{eq:4}. While GIW improves the generalization by partial (group) channel decorrelation, it strictly decorrelates neighboring channels, lacking the consideration of searching more appropriate channel combinations. It is well-known that the channels in convolution neural networks are highly related to semantics. To this end, we propose SAW module to rearrange channels, ensuring each group contains channels related to different categories. Compared with grouping same-category-related channels, decorrelation between channels from different categories is more reasonable, since different-category-related channels activate different regions. Removing the correlations between those channels not only enhances the representation capacity of each single channel, but also prevents the information loss, resulting in distributed alignment with minor changes to the semantic content.

As for segmentation model,  the segmentation results of each category are obtained by multiplying all the channels by their corresponding weights and then adding them up. The weight value determines the influence of channel on the category. Hence, to identify each channel belongs to which category, we utilize the classifier from the SAN module. As shown in \cref{fig4} (c), for each category, there are $K$ weights corresponding to $K$ channels, \ie $\{w_{\mathrm{class}~c}^{1},w_{\mathrm{class}~c}^{2},...,w_{\mathrm{class}~c}^{K}\}$ where $c\in\{1,...,C\}$. After turning them into absolute values. We rank the weights of each category from the largest to the smallest. Then in each category, the first $\frac{K}{C}$ weight indexes are selected. For the sake of clarity, we use $\mathcal{I}\in\mathbb{R}^{C\times\frac{K}{C}}$ to denote the all selected indexes, where $\mathcal{I}(i,j)$ represents the $j$-th selected index of $i$-th category, $i\in\{1,...,C\}$ and $j\in\{1,...,\frac{K}{C}\}$. We arrange $\frac{K}{C}$ groups and allocate $C$ different-category-related channels for each group. Specifically, each channel is weighted by its corresponding classifier weight before grouping, aiming to execute adaptive whitening transformation. Therefore, the $m$-th group of $n$-th sample: $\mathbf{G}_{n}^{m}$ in \cref{eq:14} can be modified as $\mathbf{\bar{G}}_{n}^{m}$:
\vspace{-0.8mm}
\begin{equation}
\begin{split}\mathbf{\bar{G}}_{n}^{m}=[\mathbf{\widetilde{F}}_{n,\mathcal{I}(1,m)}\cdot w_{\mathrm{class1}}^{\mathcal{I}(1,m)};\thinspace\mathbf{\widetilde{F}}_{n,\mathcal{I}(2,m)}\cdot w_{\mathrm{class2}}^{\mathcal{I}(2,m)};\\
...\thinspace;\thinspace\mathbf{\widetilde{F}}_{n,\mathcal{I}(C,m)}\cdot w_{\mathrm{classC}}^{\mathcal{I}(C,m)}].
\end{split}
\end{equation}

\noindent Correspondingly, our whitening loss is formulated as:
\vspace{-1.2mm}

\begin{equation}
\mathcal{L}_{\mathrm{SAW}}=\frac{1}{N}\sum_{n=1}^{N}\sum_{m=1}^{\frac{K}{C}}||\Psi(\mathbf{\bar{G}}_{n}^{m})-\mathbf{I}||_{1}.
\end{equation}

Note that operations of SAW module do not change the features of main network in forward pass. Therefore, different from SAN module, SAW is only applied during training.

% Note that operations of SAW module \ie channel selecting, weighting, grouping and covariance transforming are performed for computing the whitening loss $\mathcal{L}_{\mathrm{SAW}}$ only. These operations do not change the features of the main network in forward pass. Therefore, different from SAN module, SAW is only applied during training.

%, aiming to obtain powerful feature extractor and SAN module.

%-------------------------------------------------------------------------

\begin{table*}[t]
\caption{Performance comparison in terms of mIoU ($\%$) between Domain Generalization methods. The best and second best results are \textbf{highlighted} and \underline{underlined}, respectively. $\dagger$ denotes our re-implemention of the respective method. G, C, B, M and S denote GTA5 \cite{richter2016playing}, Cityscapes \cite{cordts2016cityscapes}, BDDS \cite{yu2018bdd100k}, Mapillary \cite{neuhold2017mapillary} and SYNTHIA \cite{ros2016synthia}, respectively.}
\vspace{-2mm}
\label{tab:DG performance}
\centering{}\resizebox{1.0\textwidth}{!}{%
\begin{tabular}{ccccccccccccccccc}
\hline 
\multirow{2}{*}{Methods} & \multirow{2}{*}{Publication} & \multirow{2}{*}{Backbone} & \multicolumn{4}{c}{Train on GTA5 (G)} &  & \multicolumn{4}{c}{Train on SYNTHIA (S)} &  & \multicolumn{4}{c}{Train on Cityscapes (C)}\tabularnewline
\cline{4-7} \cline{9-12} \cline{14-17}
 &  &  & $\rightarrow$C & $\rightarrow$B & $\rightarrow$M & $\rightarrow$S &  & $\rightarrow$C & $\rightarrow$B & $\rightarrow$M & $\rightarrow$G &  & $\rightarrow$B & $\rightarrow$M & $\rightarrow$G & $\rightarrow$S\tabularnewline
\hline 
\noalign{\vskip0.1cm}
Baseline &  &  & 28.89 & 25.44 & 26.87 & 25.72 &  & 22.90 & 23.15 & 20.81 & 25.23 &  & 43.15 & 50.91 & 41.53 & 22.36\tabularnewline
IBN$^{\dagger}$ \cite{pan2018two} & ECCV 2018 &  & 31.25 & 31.68 & 33.27 & 26.45 &  & 31.68 & 28.34 & 29.97 & 26.03 &  & 45.55 & 53.63 & 43.64 & 24.78\tabularnewline
SW \cite{pan2019switchable} & ICCV 2019 &  & 35.70 & 27.11 & 27.98 & 26.65 &  & 28.00 & 26.80 & 24.70 & 25.82 &  & 45.37 & 53.02 & 42.79 & 23.97\tabularnewline
DRPC \cite{yue2019domain} & ICCV 2019 & VGG-16 & \uline{36.11} & 31.56 & 32.25 & 26.89 &  & 35.52 & 29.45 & \uline{32.27} & 26.38 &  & 46.86 & \uline{55.83} & \uline{43.98} & 24.84\tabularnewline
GTR$^{\dagger}$ \cite{peng2021global} & TIP 2021 &  & 36.10 & 32.14 & 34.32 & 26.45 &  & 36.07 & 31.57 & 30.63 & \uline{26.93} &  & 45.93 & 54.08 & 43.72 & 24.13\tabularnewline
ISW$^{\dagger}$ \cite{choi2021robustnet} & CVPR 2021 &  & 34.36 & \uline{33.68} & \uline{34.62} & \uline{26.99} &  & \uline{36.21} & \uline{31.94} & 31.88 & 26.81 &  & \uline{47.26} & 54.21 & 42.08 & \uline{24.92}\tabularnewline
Ours &  &  & \textbf{38.21} & \textbf{36.30} & \textbf{36.87} & \textbf{28.45} &  & \textbf{38.36} & \textbf{34.32} & \textbf{33.23} & \textbf{27.94} &  & \textbf{49.19} & \textbf{56.37} & \textbf{45.73} & \textbf{26.51}\tabularnewline[0.1cm]
\hline 
\noalign{\vskip0.1cm}
Baseline &  &  & 29.32 & 25.71 & 28.33 & 26.19 &  & 23.18 & 24.50 & 21.79 & 26.34 &  & 45.17 & 51.52 & 42.58 & 24.32\tabularnewline
IBN \cite{pan2018two} & ECCV 2018 &  & 33.85 & 32.30 & 37.75 & 27.90 &  & 32.04 & 30.57 & 32.16 & 26.90 &  & 48.56 & 57.04 & 45.06 & 26.14\tabularnewline
SW \cite{pan2019switchable} & ICCV 2019 &  & 29.91 & 27.48 & 29.71 & 27.61 &  & 28.16 & 27.12 & 26.31 & 26.51 &  & 48.49 & 55.82 & 44.87 & 26.10\tabularnewline
DRPC \cite{yue2019domain} & ICCV 2019 & ResNet-50 & 37.42 & 32.14 & 34.12 & 28.06 &  & 35.65 & 31.53 & 32.74 & \uline{28.75} &  & 49.86 & 56.34 & 45.62 & \uline{26.58}\tabularnewline
GTR$^{\dagger}$ \cite{peng2021global} & TIP 2021 &  & \uline{37.53} & 33.75 & 34.52 & 28.17 &  & \uline{36.84} & \uline{32.02} & \uline{32.89} & 28.02 &  & \uline{50.75} & 57.16 & \uline{45.79} & 26.47\tabularnewline
ISW \cite{choi2021robustnet} & CVPR 2021 &  & 36.58 & \uline{35.20} & \uline{40.33} & \uline{28.30} &  & 35.83 & 31.62 & 30.84 & 27.68 &  & 50.73 & \uline{58.64} & 45.00 & 26.20\tabularnewline
Ours &  &  & \textbf{39.75} & \textbf{37.34} & \textbf{41.86} & \textbf{30.79} &  & \textbf{38.92} & \textbf{35.24} & \textbf{34.52} & \textbf{29.16} &  & \textbf{52.95} & \textbf{59.81} & \textbf{47.28} & \textbf{28.32}\tabularnewline[0.1cm]
\hline 
\noalign{\vskip0.1cm}
Baseline &  &  & 30.64 & 27.82 & 28.65 & 28.15 &  & 23.85 & 25.01 & 21.84 & 27.06 &  & 46.23 & 53.23 & 42.96 & 25.49\tabularnewline
IBN$^{\dagger}$ \cite{pan2018two} & ECCV 2018 &  & 37.42 & 38.28 & 38.28 & 28.69 &  & 34.18 & 32.63 & 36.19 & 28.15 &  & 50.22 & 58.42 & 46.33 & 27.57\tabularnewline
SW$^{\dagger}$ \cite{pan2019switchable} & ICCV 2019 &  & 36.11 & 36.56 & 32.59 & 28.43 &  & 31.60 & \uline{35.48} & 29.31 & 27.97 &  & 50.10 & 56.16 & 45.21 & 27.18\tabularnewline
DRPC \cite{yue2019domain} & ICCV 2019 & ResNet-101 & 42.53 & 38.72 & 38.05 & \uline{29.67} &  & 37.58 & 34.34 & 34.12 & \uline{29.24} &  & 51.49 & 58.62 & \uline{46.87} & 28.96\tabularnewline
GTR \cite{peng2021global} & TIP 2021 &  & \uline{43.70} & \uline{39.60} & \uline{39.10} & 29.32 &  & \uline{39.70} & 35.30 & \uline{36.40} & 28.71 &  & \uline{51.67} & 58.37 & 46.76 & \uline{29.07}\tabularnewline
ISW$^{\dagger}$ \cite{choi2021robustnet} & CVPR 2021 &  & 42.87 & 38.53 & 39.05 & 29.58 &  & 37.21 & 33.98 & 35.86 & 28.98 &  & 50.98 & \uline{59.70} & 46.28 & 28.43\tabularnewline
Ours &  &  & \textbf{45.33} & \textbf{41.18} & \textbf{40.77} & \textbf{31.84} &  & \textbf{40.87} & \textbf{35.98} & \textbf{37.26} & \textbf{30.79} &  & \textbf{54.73} & \textbf{61.27} & \textbf{48.83} & \textbf{30.17}\tabularnewline[0.1cm]
\hline 
\end{tabular}}
\vspace{-1mm}
\end{table*}

%--------------------------figure 4
    \begin{figure*}[t]
    \centering{}
     \includegraphics[scale=0.652]{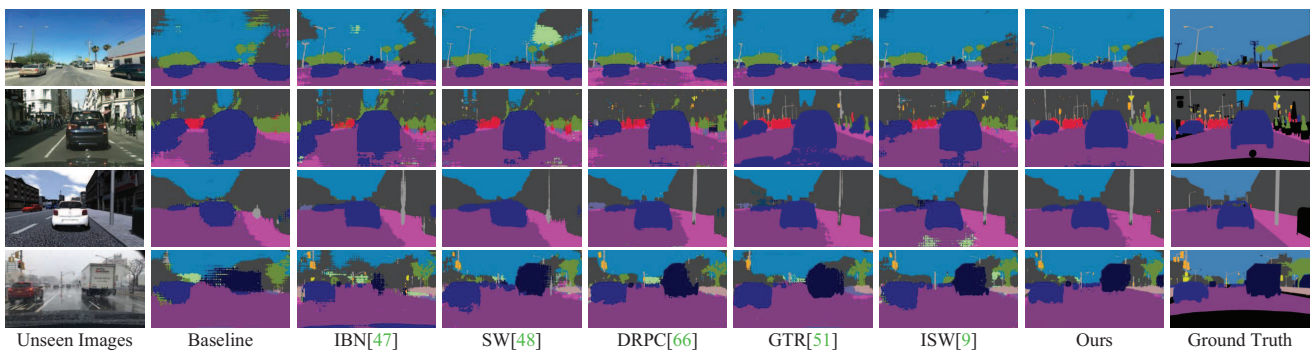} 
     \vspace{-6mm}
     \caption{Visual comparison with different Domain Generalization methods on unseen domains \ie Cityscapes \cite{cordts2016cityscapes}, BDDS \cite{yu2018bdd100k}, Mapillary \cite{neuhold2017mapillary} and SYNTHIA \cite{ros2016synthia}, with the model trained on GTA5 \cite{richter2016playing}. The backbone network is ResNet-50.}
    \label{fig5}\vspace{-5mm}
    
    \end{figure*}
%--------------------------figure 4

\section{Experiments}

\subsection{Datasets Description}
\textbf{Synthetic Datasets.} {GTA5} \cite{richter2016playing} is a synthetic image dataset, which is  collected by using  GTA-V game engine. It contains 24966 images with a resolution of 1914 $\times$ 1052 along with their pixel-wise semantic labels. 
% It includes 19 classes which are compatible with other semantic segmentation datasets. 
{SYNTHIA} \cite{ros2016synthia} is a large synthetic dataset with pixel-level semantic annotations. The subset {SYNTHIA-RANDCITYSCAPES} \cite{ros2016synthia} is used in our experiments which contains 9400 images with a high resolution of 1280 $\times$ 760.

\textbf{Real-World Datasets.} {Cityscapes} \cite{cordts2016cityscapes} is a high resolution dataset (\ie 2048 $\times$ 1024) of 5000 vehicle-captured urban street images taken from from 50 different cities primarily in Germany. {BDDS} \cite{yu2018bdd100k} contains thousands of real-world dashcam video frames with accurate pixel-wise annotations, where 10000 images are provided with a resolution of 1280 $\times$ 720. {Mapillary} \cite{neuhold2017mapillary} contains 25000 images with diverse resolutions. The annotations contain 66 object categories, but only 19 categories overlap with others.

\subsection{Implementation Details \label{implementation}}

We initialize the weights of the feature extractor module with an ImageNet \cite{deng2009imagenet} pre-trained model. We use SGD \cite{krizhevsky2012imagenet} optimizer with with an initial learning rate of 5e-4, a batch size of 2, a momentum of 0.9 and a weight decay of 5e-4. Besides, we follow the polynomial learning rate scheduling \cite{chen2017rethinking} with the power of 0.9. We train model for 200000 iterations. All datasets have 19 common categories, thus the parameter $C$ defined in both SAN and SAW is set to 19. However, since SAW arranges $\frac{K}{C}$ groups, $C$ can only be 2, 4, 8 or 16 to make $K$ (channel number) divisible by $C$. Based on the results of ablation study on parameter $C$ (\cref{sec: hyper-param}
), we set $C=4$ in both SAN and SAW.

We implement our method on PyTorch \cite{paszke2017automatic} and use a single NVIDIA RTX 3090 GPU for our experiments. %\MH{this sentence isnt clear Also, as suggested by \cite{pan2018two,choi2021robustnet}, we embed our modules after the first two convolution groups.} 
% See \textcolor{red}{Appendix A} for further implementation details.
Following previous works, we use PASCAL VOC Intersection over Union (IoU) \cite{everingham2015pascal} as the evaluation metric.%, where mIoU is the mean value of IoUs computed across all categories.

\subsection{Comparison with DG and DA methods}
For brevity, we use G, S, C, B and M to denote GTA5 \cite{richter2016playing}, SYNTHIA \cite{ros2016synthia}, Cityscapes \cite{cordts2016cityscapes}, BDDS \cite{yu2018bdd100k} and Mapillary \cite{neuhold2017mapillary}, respectively. 
% Extensive experiments are conducted to show the generalization performance of our framework. 
We extensively evaluate our method with different backbones including VGG-16 \cite{simonyan2014very}, ResNet-50 and ResNet-101 \cite{he2016deep}. We repeat each experiment three times, and report the average results.
% \textbf{Moreover, in this subsection, we perform each experiment three times and average for fair comparisons.} 
Unlike existing synthetic-to-real generalization approaches, 
% Different from previous work mainly focusing on synthetic-to-real generalization, 
we propose to evaluate our model from an arbitrary domain to other unseen domains. Therefore, we conduct comprehensive experiments on five generalization settings, from \textbf{(1)} G to C, B, M \& S; \textbf{(2)} S to C, B, M \& G; \textbf{(3)} C to G, S, B \& M; \textbf{(4)} B to G, S, C \& M; \textbf{(5)} M to G, S, C \& B.
Our results reported in \cref{tab:DG performance} on the first three settings, and \textcolor{red}{Appendix A} on the remaining 2 settings, suggest that our model consistently gains the best performance across all settings and backbones.
%Due to lack of space, we only present the experimental results of the first three settings in Tab. \ref{tab:DG performance}, because existing DG methods mainly compare results under these three settings. 
% The results of the other two settings are provided in the supplementary Section \textcolor{red}{A}, which show a similar performance trend. 
% As shown in Tab. \ref{tab:DG performance}, our model consistently gains the best performance on all settings and backbones. 
Compared to the the second best results (see \underline{underlined} values), our method shows a large improvement. Visual samples for qualitative comparison are given in \cref{fig5}. Remarkably, our method performs favorably in comparison to the methods that have access to the target domain data, see results in \textcolor{red}{Appendix B}.  

\subsection{Source Domain Performance Decay Analysis}

A common pitfall of the domain generalization methods is that their performance degrades on the source domain. To compare different methods for this aspect, we evaluate them on the test set of the source domain in \cref{Tab: source decay}. The results suggest that our method largely retains performance on the source domain, and performs comparably with the model trained without domain-generalization (Baseline in \cref{Tab: source decay}). %The results in \cref{Tab: source decay} are with ResNet-50 backbone. %For results on other backbones, which show same performance trend, see \textcolor{red}{Appendix A}. \MH{You do not need to refer to all experiments in the suppl. in the main paper. I think we can skip this one.}

% Compared to baseline, model with DG technique can well generalize to other domains but fall into the problem of performance degradation on source domain. To prove our method is capable to achieve generalization with negligible source domain performance drop, we utilize our trained DG models, which perform well on other unseen domains, to directly evaluate on the test set of source domain. As shown in Tab. \ref{Tab: source decay}, compared to the model without DG technique (i.e., baseline), our approach performs a slight performance decay (within 0.5$\%$) while other DG methods dampen the performance by 1$\%$$\sim$3$\%$. we only present the performance comparison on VGG-16, results on other backbones, which show same performance trend, are listed in supplementary Section \textcolor{red}{A}.

\subsection{Ablation Study}

% \subsubsection{Effectiveness of SAN and SAW}
\textbf{SAN and SAW.} We investigate the individual contribution of SAN and SAW modules towards overall performance. \cref{tab:ablation-on-module} shows the mIoU improvement on ResNet-50 once we progressively integrate SAN and SAW. Experiments are conducted for generalization from GTA5 (G) to the other four datasets \ie Cityscapes (C), BDDS (B), Mapillary (M) and SYNNTHIA(S). In our case, each of them helps boost the generalization performance by a large margin. Specifically, we observe that SAN and SAW greatly achieve an average improvement of 8.71$\%$ and 7.93$\%$, respectively. We further observe that the models with only SAW show slightly weaker generalization capacity than those with only SAN. This is because without category-level centering of SAN, SAW performs distributed alignment, which may lead to incorrect feature matching between different categories. Therefore, our SAW module is mainly proposed to complement SAN in an integrated approach. As shown in \cref{tab:ablation-on-module}, the best performance is achieved with a combination of both SAN and SAW. 

\textbf{CFR block in SAN.} To demonstrate the effectiveness of the Category-level Feature Refinement (CFR) in SAN, we conducted an ablation experiment by removing the CFR block. As shown in \cref{tab: CFR and ClassGrouping} (top row), model without CFR consistently performs worse than model with SAN. With the help of CFR, SAN achieves an average gain of 2.30$\%$, which demonstrates its usefulness. 

\textbf{Category-related Grouping in SAW.} We perform ablations on different grouping strategies to verify their contributions. We conduct experiments on baseline with IW, GIW and SAW, respectively. The architectures of these three models are illustrated in \cref{fig4}. As shown in \cref{tab: CFR and ClassGrouping} (bottom row), by adopting the general grouping operation, GIW shows significant improvement compared to model with IW. When applying our proposed SAW which performs category-related grouping, the performance of network improves to 37.54$\%$, 34.97$\%$, 39.85$\%$ \& 28.46$\%$ on G $\rightarrow$ C, B, M \& S. This confirms the effectiveness of category-related grouping in SAW. 

\begin{figure}[t]
    \centering{}\vspace{0mm}
     \includegraphics[scale=0.55]{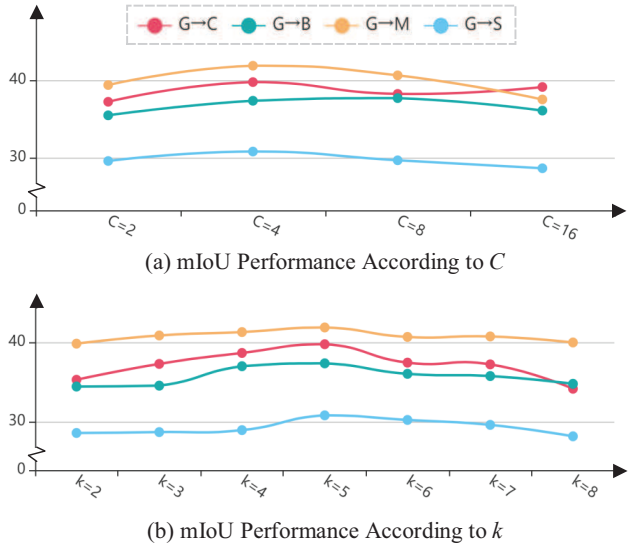} 
     \vspace{-6mm}
     \caption{Change in performance with hyper-parameters: $C$ and $k$. The experimental backbone is ResNet-50.
     }
    \label{fig6}\vspace{-5mm}
    
    \end{figure}

\textbf{Hyper-Parameter Analysis.\label{sec: hyper-param}} $C$ decides on the number of categories for semantic-aware feature alignment in both SAN and SAW. Since we rank the categories according to the their respective proportions in training data, only the first $C$ categories are selected to handle category-level feature matching. To ensure that the channel dimension of feature maps is divisible by $C$, we can only set $C\in\{2,4,6,8,16\}$. \cref{fig6} (a) shows compares results for different values, suggesting the optimal $C$ is 4. In CFR block, we adopt $k$-means clustering to separate feature elements into category region and background. For searching the optimal parameter $k$, we only choose the first cluster whose cluster center is highest as the category region. As shown in \cref{fig6} (b), model performs best when adopting $k$=5.
\vspace{-2.75mm}

%--------------------------figure 5
%--------------------------figure 5

% , we study how each module in our approach influences the overall performance. The experimental setting is the generalization from GTA5 (G) to the other four datasets, i.e., Cityscapes (C), BDDS (B), Mapillary (M) and SYNNTHIA(S). Extensive ablations are conducted on VGG-16, ResNet-50 and ResNet-101. Tab. \ref{tab:ablation-on-module} details the mIoU improvement by progressively adding the proposed modules: Semantic-Aware Normalization (SAN) and Semantic-Aware Whitening (SAW). SAN and SAW are generic ways to alleviate domain shift. In our case, each of them helps boost the generalization performance in a large margin, where SAN and SAW greatly achieve an average improvement of 8.13$\%$ and 6.85$\%$, respectively. It can be seen that models with only SAW show slightly weaker generalization capacity than those with only SAN. This is because without category-level centering of SAN, SAW executes distributed uniformization, may leading to incorrect feature matching between different categories. Therefore, our SAW module is mainly proposed to complement with SAN for an integrated approach. As shown in Tab. \ref{tab:ablation-on-module}, the combination of SAN and SAW achieves remarkable results and gives a further significant improvement on all generalization settings.

\begin{table}[t]
\caption{Comparison of different methods for performance drop on source domain. The performance drop ($\Downarrow$) is obtained with respect to the baseline. The best and second best values are \textbf{highlighted} and \underline{underlined}, respectively. The network backbone is ResNet-50.\label{Tab: source decay}}
\vspace{-2mm}
\centering{}\resizebox{0.47\textwidth}{!}{%
\begin{tabular}{cccccc}
\toprule[0.2em]
% \hline 
\multirow{1}{*}{Methods} & \multirow{1}{*}{GTA5} & \multirow{1}{*}{SYNTHIA} & Cityscapes & BDDS & Mapillary\tabularnewline
\hline 
Baseline & 73.95 & 70.84 & 77.93 & 79.67 & 70.49\tabularnewline
\midrule[0.15em] 
\noalign{\vskip0.1cm}
IBN$^{\dagger}$ \cite{pan2018two} & $\Downarrow$1.05 & $\Downarrow$2.40 & $\Downarrow$1.38 & $\Downarrow$\uline{1.30} & $\Downarrow$2.25\tabularnewline
SW \cite{pan2019switchable} & $\Downarrow$\uline{0.45} & $\Downarrow$1.71 & $\Downarrow$\uline{0.63} & $\Downarrow$1.52 & $\Downarrow$\uline{0.91}\tabularnewline
DRPC \cite{yue2019domain} & $\Downarrow$2.37 & $\Downarrow$3.63 & $\Downarrow$2.72 & $\Downarrow$2.46 & $\Downarrow$3.05\tabularnewline
GTR$^{\dagger}$ \cite{peng2021global} & $\Downarrow$2.07 & $\Downarrow$2.11 & $\Downarrow$3.25 & $\Downarrow$3.64 & $\Downarrow$4.17\tabularnewline
ISW$^{\dagger}$ \cite{choi2021robustnet} & $\Downarrow$1.85 & $\Downarrow$\uline{1.28} & $\Downarrow$1.52 & $\Downarrow$1.83 & $\Downarrow$1.23\tabularnewline
Ours & $\Downarrow$\textbf{0.19} & $\Downarrow$\textbf{0.08} & $\Downarrow$\textbf{0.06} & $\Downarrow$\textbf{0.23} & $\Downarrow$\textbf{0.34}\tabularnewline[0.1cm]
\bottomrule[0.15em] 
\end{tabular}}
\vspace{-2mm}
\end{table}

\begin{table}[t]
\caption{Ablation analysis of SAN (\cref{subsec:Spatial Feature Constraint (SFC)}) and SAW (\cref{subsec:Channel Feature Constraint (CFC)}).} %The network backbone is ResNet-50.}
\vspace{-2mm}
\label{tab:ablation-on-module}
\centering{}\doublerulesep=0.5pt \resizebox{0.47\textwidth}{!}{%
\begin{tabular}{ccccccc}
\toprule[0.2em]
\multirow{2}{*}{Methods} & \multirow{2}{*}{SAN} & \multirow{2}{*}{SAW} & \multicolumn{4}{c}{Train on GTA5 (G)}\tabularnewline
\cline{4-7}
 &  &  & C & B & M & S\tabularnewline
\midrule[0.15em] 
\noalign{\vskip0.1cm}
Baseline &  &  & 29.32 & 25.71 & 28.33 & 26.19\tabularnewline
+ SAN & $\checked$ &  & 38.92 & 36.43 & 40.11 & 28.91\tabularnewline
+ SAW &  & $\checked$ & 37.54 & 34.97 & 39.85 & 28.46\tabularnewline
All & $\checked$ & $\checked$ & \textbf{39.75} & \textbf{37.34} & \textbf{41.86} & \textbf{30.79}\tabularnewline[0.1cm]
\bottomrule[0.15em] 
\end{tabular}}
\vspace{-2mm}
\end{table}

\begin{table}[t]
\caption{Ablation of different blocks in SAN and SAW}%. Bottom: ablation on Class-related Grouping in SAW. The backbone is ResNet-50.}
\vspace{-2mm}
\label{tab: CFR and ClassGrouping}
\centering{}\doublerulesep=0.5pt \resizebox{0.47\textwidth}{!}{%
\begin{tabular}{ccccc}
\toprule[0.2em] 
\multirow{2}{*}{Methods} & \multicolumn{4}{c}{Train on GTA5 (G)}\tabularnewline
\cline{2-5} 
 & C & B & M & S\tabularnewline
\midrule[0.15em] 
Baseline & 29.32 & 25.71 & 28.33 & 26.19\tabularnewline
Baseline + SAN (w/o CFR) & 35.51 & 33.24 & 38.68 & 27.76\tabularnewline
Baseline + SAN & \textbf{38.92} & \textbf{36.43} & \textbf{40.11} & \textbf{28.91}\tabularnewline
\midrule[0.15em] 
Baseline + IW & 33.19 & 31.27 & 30.55 & 26.55\tabularnewline
Baseline + GIW & 35.76 & 32.88 & 36.95 & 27.24\tabularnewline
Baseline + SAW & \textbf{37.54} & \textbf{34.97} & \textbf{39.85} & \textbf{28.46}\tabularnewline
\bottomrule[0.15em] 
\end{tabular}}
\vspace{-4mm}
\end{table}

\section{Conclusion and limitations}

In this manuscript, we present a domain generalization approach to address the out-of-domain generalization for semantic segmentation. We propose two novel modules: Semantic-Aware Normalization (SAN) and Semantic-Aware Whitening (SAW), which sequentially perform category-level center alignment and distributed alignment to achieve both domain-invariant and discriminative features. Comprehensive experiments demonstrate the effectiveness of SAN and SAW with state-of-the-art performance in both domain generalization and domain adaptation. Although the method effectively eliminates feature differences caused by style variations on source domain, the extracted style-invariant features may still contain source-domain specific cues, leading to significant performance delta between the source and target domain. Some interesting future directions to close this gap include learning how to model domain shift (meta-learning), integrating multiple domain-specific neural networks (ensemble learning) and teaching a model to perceive generic features regardless of the target task (disentangled representation learning).

% Although the method effectively eliminates feature differences caused by style variations on source domain, the extracted style-invariant features may still contain source-domain specific cues. Therefore in our future work, we consider to further investigate feature extraction of domain-invariant content such as shape and spatial layout while combining with our semantic-aware strategy.

\textit{The datasets used in the paper lack diversity and have biases as they are mostly captured in the developed world. Beyond that, this paper have no ethical problem including personally identifiable information, human subject experimentation and military application.} %}Beyond that, this paper have no ethical problem including personally identifiable information, human subject experimentation and military application.}

\noindent \textbf{Acknowledgement:} This work was partially supported by the National Natural Science Foundation of China (No. 61972435, U20A20185). M.~Hayat is supported by ARC DECRA fellowship DE200101100 .

\balance
%%%%%%%%% REFERENCES
{\small
\bibliographystyle{ieee_fullname}
\bibliography{egbib}
}

\end{document}


%
\setcounter{section}{0}
\setcounter{table}{0}
\setcounter{figure}{0}
\def\thesection{Appendix \Alph{section}}
% \twocolumn[
%   \begin{@twocolumnfalse}
%     \begin{center}
%         \section*{Supplementary: Semantic-Aware Domain Generalized Segmentation}
%     \end{center}
%   \end{@twocolumnfalse}
% ]

% In this supplementary, additional information is provided as following five aspects:

% \textbf{A.} comprehensive evaluation on other DG settings;

% \textbf{B.} comparison with Domain Adaptation methods;

% \textbf{C.} detailed framework structure and implementations;

% \textbf{D.} discussion on computational complexity;

% \textbf{E.} additional visualization of semantic segmentation.

\vspace{2mm}
\section{Evaluation on other DG settings \label{Appen: DG compare}}
\vspace{0mm}

Existing DG methods \cite{pan2018two,pan2019switchable,yue2019domain,choi2021robustnet,peng2021global} only focus on three domain generalization settings, \ie (1) G $\rightarrow$ C, B, M, \& S; (2) S $\rightarrow$ C, B, M, \& G and (3) C $\rightarrow$ G, S, B, \& M while losing the sight of the other two DG settings: (4) B $\rightarrow$ G, S, C, \& M and (5) M $\rightarrow$ G, S, C, \& M. However, recently several studies \cite{hoffman2016fcns,chen2017no} stress the importance of the last two settings. Therefore, we consider a more comprehensive evaluation which performs generalization from each of them. Results on the last two settings are reported in \cref{tab: DG_other}, suggesting that our model consistently achieves the state-of-the-art results on all settings and backbones.

% training from GTA5 \cite{richter2016playing}, SYNTHIA \cite{ros2016synthia} and Cityscapes \cite{cordts2016cityscapes} respectively then generalizing to other domains. 
% But such 3 DG setting lacks the diversity of 

% a comparison with DA
% methods under the same setting is impossible

% In Table 5a and Table 7 of the main paper, we show the
% quantitative comparison, through the mIoU, between our
% approach and other methods, on the 2D and cross-modal semantic segmentation benchmark. Correspondingly, we here
% provide more detailed experimental results in Table S5 and
% Table S6, covering the per-class IoU results.

% In the main paper, we represent our quantitative results compared to other Domain Generalization (DG) methods on 3 main settings: \textbf{(1)} G $\rightarrow$ C, B, M \& S, \textbf{(2)} S $\rightarrow$ C, B, M \& G and \textbf{(3)} C $\rightarrow$ G, S, B \& M.  To further verify the effectiveness of our method, we evaluate our approach on other two generalization settings, where existing DG methods are rarely compared. Specifically, we tested generalization ability of our model on the other two DG settings: \textbf{(1)} B $\rightarrow$ G, S, C \& M and \textbf{(2)} M $\rightarrow$ G, S, C \& B, with 3 different backbone networks, VGG-16, ResNet-50 and ResNet-101.
% As shown in \cref{tab: DG_other}, we can see that our proposed framework consistently achieves superior results on all settings and backbones, which comprehensively demonstrate the effectiveness of our method.

% Table 1
%----------------------------------------------
\begin{table}[th]

\caption{Performance comparison in terms of mIoU ($\%$) between DG methods. The best and second best results are \textbf{highlighted} and \underline{underlined}, respectively. $\dagger$ denotes our re-implemention of the respective method. G, C, B, M and S denote GTA5, Cityscapes, BDDS, Mapillary and SYNTHIA, respectively.\label{tab: DG_other}}
\vspace{-0mm}
\centering{}\resizebox{0.47\textwidth}{!}{%
\begin{tabular}{ccccccccccc}
\toprule[0.2em]
\multirow{2}{*}{Methods} & \multirow{2}{*}{Backbone} & \multicolumn{4}{c}{Train on BDDS (B)} &  & \multicolumn{4}{c}{Train on Mapillary (M)}\tabularnewline
\cline{3-6} \cline{8-11} 
 &  & $\rightarrow$G & $\rightarrow$S & $\rightarrow$C & $\rightarrow$M &  & $\rightarrow$G & $\rightarrow$S & $\rightarrow$C & $\rightarrow$B\tabularnewline
\midrule[0.15em] 
\noalign{\vskip0.1cm}
Baseline &  & 25.30 & 21.08 & 38.76 & 23.48 &  & 25.34 & 22.16 & 36.13 & 24.17\tabularnewline
IBN$^{\dagger}$ \cite{pan2018two} &  & 29.47 & 26.40 & 39.72 & 26.12 &  & 29.68 & 26.31 & 41.39 & 29.48\tabularnewline
SW$^{\dagger}$ \cite{pan2019switchable} &  & 27.10 & 25.23 & 39.54 & 25.67 &  & 28.70 & 25.57 & 39.66 & 28.37\tabularnewline
DRPC$^{\dagger}$ \cite{yue2019domain} & VGG-16 & \uline{32.83} & 28.06 & 40.17 & 29.00 &  & 31.53 & 28.03 & 45.15 & 30.46\tabularnewline
GTR$^{\dagger}$ \cite{peng2021global} &  & 32.75 & 27.63 & 41.06 & 29.71 &  & \uline{32.67} & 27.32 & 44.47 & 31.83\tabularnewline
ISW$^{\dagger}$ \cite{choi2021robustnet} &  & 32.60 & \uline{28.58} & \uline{42.21} & \uline{30.54} &  & 32.65 & \uline{28.44} & \uline{45.68} & \uline{32.06}\tabularnewline
Ours &  & \textbf{34.11} & \textbf{30.13} & \textbf{44.62} & \textbf{32.06} &  & \textbf{33.86} & \textbf{30.67} & \textbf{47.51} & \textbf{33.07}\tabularnewline[0.1cm]
\bottomrule[0.15em] 
\noalign{\vskip0.1cm}
Baseline &  & 26.12 & 21.65 & 39.03 & 23.87 &  & 25.46 & 23.41 & 36.79 & 26.37\tabularnewline
IBN$^{\dagger}$ \cite{pan2018two} &  & 28.97 & 25.42 & 41.06 & 26.56 &  & 30.68 & 27.01 & 42.77 & 31.01\tabularnewline
SW$^{\dagger}$ \cite{pan2019switchable} &  & 27.68 & 25.37 & 40.88 & 25.83 &  & 28.47 & 27.43 & 40.69 & 30.54\tabularnewline
DRPC$^{\dagger}$ \cite{yue2019domain} & ResNet-50 & 33.19 & 29.77 & 41.30 & \uline{31.86} &  & 33.04 & 29.59 & 46.21 & \uline{32.92}\tabularnewline
GTR$^{\dagger}$ \cite{peng2021global} &  & \uline{33.25} & \uline{30.61} & 42.58 & 30.73 &  & 32.86 & \uline{30.26} & 45.84 & 32.63\tabularnewline
ISW$^{\dagger}$ \cite{choi2021robustnet} &  & 32.74 & 30.53 & \uline{43.50} & 31.57 &  & \uline{33.37} & 30.15 & \uline{46.43} & 32.57\tabularnewline
Ours &  & \textbf{34.75} & \textbf{31.84} & \textbf{44.94} & \textbf{33.21} &  & \textbf{34.01} & \textbf{31.55} & \textbf{48.65} & \textbf{34.62}\tabularnewline[0.1cm]
\bottomrule[0.15em] 
\noalign{\vskip0.1cm}
Baseline &  & 25.84 & 24.62 & 42.06 & 24.70 &  & 26.81 & 23.74 & 39.68 & 27.19\tabularnewline
IBN$^{\dagger}$ \cite{pan2018two} &  & 30.28 & 29.06 & 44.92 & 29.90 &  & 32.07 & 28.83 & 44.89 & 30.27\tabularnewline
SW$^{\dagger}$ \cite{pan2019switchable} &  & 28.34 & 26.74 & 44.28 & 27.58 &  & 30.31 & 24.06 & 42.33 & 28.65\tabularnewline
DRPC$^{\dagger}$ \cite{yue2019domain} & ResNet-101 & 34.13 & 31.75 & \uline{46.73} & 32.63 &  & \uline{36.40} & 30.27 & 46.16 & 32.17\tabularnewline
GTR$^{\dagger}$ \cite{peng2021global} &  & \uline{35.26} & 31.98 & 45.34 & 33.27 &  & 34.65 & 29.56 & 47.68 & 33.98\tabularnewline
ISW$^{\dagger}$ \cite{choi2021robustnet} &  & 34.87 & \uline{32.89} & 46.15 & \uline{34.17} &  & 35.53 & \uline{30.92} & \uline{48.54} & \uline{34.02}\tabularnewline
Ours &  & \textbf{37.56} & \textbf{33.83} & \textbf{48.32} & \textbf{35.24} &  & \textbf{37.72} & \textbf{32.63} & \textbf{50.07} & \textbf{35.79}\tabularnewline[0.1cm]
\bottomrule[0.15em] 
\end{tabular}}
\vspace{-0mm}
\end{table}
%----------------------------------------------

\section{Comparison with DA methods}

% Table 2
%----------------------------------------------
\begin{table}[t]
\caption{Comparison results between ours and Domain Adaptation methods on GTA5$\rightarrow$Cityscapes. DA and DG denote Domain Adaption and Domain Generalization respectively. \label{tab:DA_G_to_C}}
\vspace{-0mm}
\centering{}\doublerulesep=0.5pt \resizebox{0.40\textwidth}{!}{
\begin{tabular}{ccccc}
\toprule[0.2em]
Backbone & Task & Method & Access Tgt & mIoU\tabularnewline
\midrule[0.15em] 
\multirow{15}{*}{VGG-16} & \multirow{14}{*}{DA} & FCN wild \cite{hoffman2016fcns} & $\checked$ & 27.1\tabularnewline
 &  & CDA \cite{zhang2017curriculum} & $\checked$ & 28.9\tabularnewline
 &  & CyCADA \cite{hoffman2018cycada} & $\checked$ & 34.8\tabularnewline
 &  & ROAD \cite{chen2018road} & $\checked$ & 35.9\tabularnewline
 &  & I2I \cite{murez2018image} & $\checked$ & 31.8\tabularnewline
 &  & AdaptSegNet \cite{tsai2018learning} & $\checked$ & 35.0\tabularnewline
 &  & SSF-DAN \cite{du2019ssf} & $\checked$ & 37.7\tabularnewline
 &  & DCAN \cite{wu2018dcan} & $\checked$ & 36.2\tabularnewline
 &  & CBST \cite{zou2018domain} & $\checked$ & 30.9\tabularnewline
 &  & CLAN \cite{luo2019taking} & $\checked$ & 36.6\tabularnewline
 &  & ADVENT \cite{vu2019advent} & $\checked$ & 36.1\tabularnewline
 &  & DPR \cite{tsai2019domain} & $\checked$ & 37.5\tabularnewline
 &  & BDL \cite{li2019bidirectional} & $\checked$ & \uline{41.3}\tabularnewline
 &  & FDA \cite{yang2020fda} & $\checked$ & \textbf{42.2}\tabularnewline
\cline{2-5}
 & DG & Ours & $\times$ & 38.2\tabularnewline
\midrule[0.15em] 
\multirow{11}{*}{Resnet-101} & \multirow{10}{*}{DA} & CyCADA \cite{hoffman2018cycada} & $\checked$ & 42.7\tabularnewline
 &  & ROAD \cite{chen2018road} & $\checked$ & 39.4\tabularnewline
 &  & I2I \cite{murez2018image} & $\checked$ & 35.4\tabularnewline
 &  & AdaptSegNet \cite{tsai2018learning} & $\checked$ & 41.4\tabularnewline
 &  & DCAN \cite{wu2018dcan} & $\checked$ & 41.7\tabularnewline
 &  & CLAN \cite{luo2019taking} & $\checked$ & 43.2\tabularnewline
 &  & ADVENT \cite{vu2019advent} & $\checked$ & 43.8\tabularnewline
 &  & DPR \cite{tsai2019domain} & $\checked$ & \uline{46.5}\tabularnewline
 &  & IntraDA \cite{pan2020unsupervised} & $\checked$ & 46.3\tabularnewline
 &  & DADA \cite{vu2019dada} & $\checked$ & \textbf{47.3}\tabularnewline
\cline{2-5}
 & DG & Ours & $\times$ & 45.3\tabularnewline
\bottomrule[0.15em] 
\end{tabular}}
\vspace{-0mm}
\end{table}
%----------------------------------------------

% Table 3
%----------------------------------------------
\begin{table}[t]

\caption{Comparison results between ours and Domain Adaptation methods on SYNTHIA$\rightarrow$Cityscapes. \label{tab:DA_S_to_C}}
\vspace{-0mm}
\centering{}\doublerulesep=0.5pt \resizebox{0.40\textwidth}{!}{%
\begin{tabular}{ccccc}
\toprule[0.2em]
Backbone & Task & Method & Access Tgt & mIoU\tabularnewline
\midrule[0.15em] 
\multirow{10}{*}{VGG-16} & \multirow{9}{*}{DA} & FCN wild \cite{hoffman2016fcns} & $\checked$ & 20.2\tabularnewline
 &  & CDA \cite{zhang2017curriculum} & $\checked$ & 29.0\tabularnewline
 &  & ROAD \cite{chen2018road} & $\checked$ & 36.2\tabularnewline
 &  & DCAN \cite{wu2018dcan} & $\checked$ & 35.4\tabularnewline
 &  & CBST \cite{zou2018domain} & $\checked$ & 35.4\tabularnewline
 &  & ADVENT \cite{vu2019advent} & $\checked$ & 31.4\tabularnewline
 &  & DPR \cite{tsai2019domain} & $\checked$ & 33.7\tabularnewline
 &  & BDL \cite{li2019bidirectional} & $\checked$ & \uline{39.0}\tabularnewline
 &  & FDA \cite{yang2020fda} & $\checked$ & \textbf{40.5}\tabularnewline
\cline{2-5}
 & DG & Ours & $\times$ & 37.4\tabularnewline
\midrule[0.15em] 
\multirow{5}{*}{Resnet-101} & \multirow{4}{*}{DA} & ADVENT \cite{vu2019advent} & $\checked$ & 40.8\tabularnewline
 &  & DPR \cite{tsai2019domain} & $\checked$ & 40.0\tabularnewline
 &  & IntraDA\cite{pan2020unsupervised} & $\checked$ & \uline{41.7}\tabularnewline
 &  & DADA \cite{vu2019dada} & $\checked$ & \textbf{42.6}\tabularnewline
\cline{2-5}
 & DG & Ours & $\times$ & 40.9\tabularnewline
\bottomrule[0.15em]
\end{tabular}}
\vspace{-2mm}
\end{table}
%----------------------------------------------

% All comparisons in the main paper are conducted with Domain Generalization (DG) manner, where the network training has no access to target domain. Now we compare the result of our method with those reported from several domain adaptation (DA) methods which are target domain-accessible. In view of the most of existing DA methods represent adaptation results from GTA5/SYNTHIA to Cityscapes with backbones VGG-16 and ResNet-101,  we report the comparison results under the same settings in \cref{tab:DA_G_to_C} and \cref{tab:DA_S_to_C}. Although our method may not the best performer, it shows on par or even better performance than existing DA methods.

Domain Adaptation (DA) methods require access to the target domain to solve domain shift problems. In contrast, our method is designed in Domain Generalization (DG) manner for broad generalization to totally unseen domains without accessing any target domain data.
Therefore, the target domain-accessible DA methods have the inherent performance superiority than DG methods which are target domain-agnostic.
In order to see whether our approach is up to the performance standard of DA, we compare the results of our method with those reported from several previous state-of-the-art DA methods.
From \cref{tab:DA_G_to_C} and \cref{tab:DA_S_to_C}, we can see that the generalization performance of our method outperforms the adaptation performance of most other techniques. In addition, no target-domain data is needed in our method, resulting in more extensive applicability.

\section{Further Implementation Details}

\begin{figure}[h]
    \centering{}\vspace{-0mm}
     \includegraphics[scale=0.38]{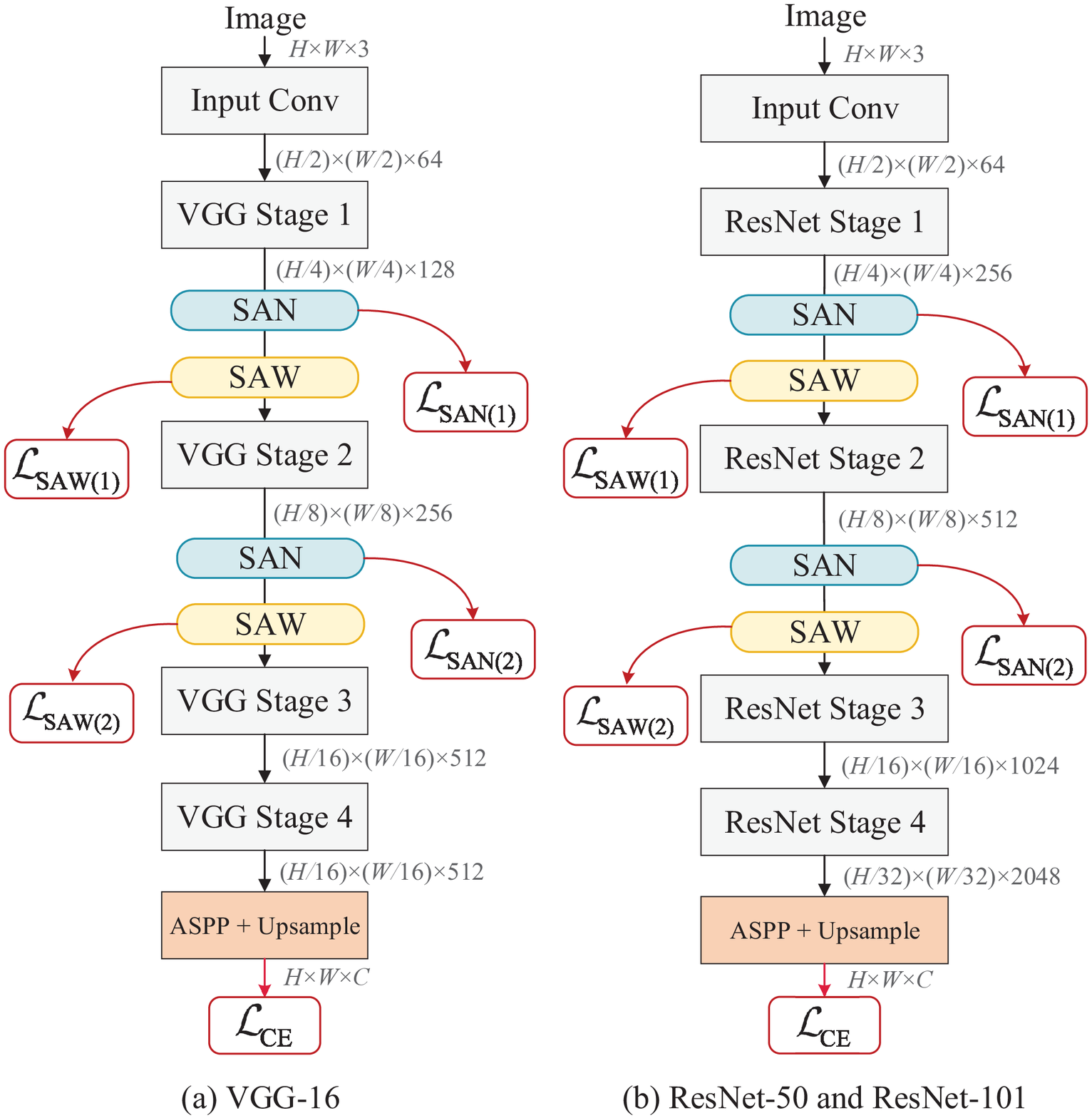} 
     \caption{Detailed Architecture of our approach with the backbone of VGG and ResNet. \label{fig: detailed architecture}
     }
     \vspace{-2mm}
    
    \end{figure}

We follow previous work \cite{pan2018two,choi2021robustnet} to adopt normalization and whitening at the first two stages of convolution layers, since shallow layers encode more style information \cite{pan2018two}. As shown in \cref{fig: detailed architecture}, for each backbone network, we impose SAN and SAW after stage 1 and stage 2.
% In the end, the final loss is totally formulated as:
% \[
% \begin{aligned}\mathcal{L}=\mathcal{L}_{\mathrm{CE}}+\alpha_{1}(\mathcal{L}_{\mathrm{SAN(1)}}+\mathcal{L}_{\mathrm{SAN(2)}})\\
% +\alpha_{2}(\mathcal{L}_{\mathrm{SAW(1})}+\mathcal{L}_{\mathrm{SAW(2)}}),
% \end{aligned}
% \]
% where $\alpha_{1}$ and $\alpha_{2}$ are set to 0.2 and 0.8 respectively, $\mathcal{L}_{\mathrm{CE}}$ is the Cross Entropy Loss for segmentation training, $\mathcal{L}_{\mathrm{SAN(1)}}$ and $\mathcal{L}_{\mathrm{SAW(1)}}$ are losses within SAN and SAW after stage 1, while $\mathcal{L}_{\mathrm{SAN(2)}}$ and $\mathcal{L}_{\mathrm{SAW(2)}}$ are those after stage 2.

%--------------------------figure 1
%--------------------------figure 1

% \subsection{SAN and SAW.}

% \cref{tab: san_saw_other} shows the ablation results on SAN and SAW with other two backbones. It can be seen that the performance improves significantly along with the progressive addition of the proposed two modules (SAN and SAW), which demonstrates the complementarity in them. Besides, each of them largely boost the generalization performance individually demonstrating the effectiveness of each module.

%----------------------------------------------

\section{Computational complexity}

As shown in \cref{tab:compute}, compared to the baseline, our methods performs domain generalization with negligible addition in both training and inference time. This is because the proposed modules are only implemented in the first two layers of the network, for only four main categories \textit{i.e.} $C=4$, see Sec. \textcolor{red}{5.5} of the main paper. The additional memory overhead from our modules is less than 2G. 

\begin{table}[h]
\vspace{0mm}
\caption{Comparison on computation cost. \label{tab:compute}}
\vspace{-1mm}
\centering{}\doublerulesep=0.5pt \resizebox{0.47\textwidth}{!}{%
\begin{tabular}{ccccc}
\toprule[0.2em]
Backbone & Methods & Memory (G) & Training Time (s) & Inference Time (ms)\tabularnewline
\midrule[0.15em] 
\multirow{2}{*}{Vgg-16} & Baseline & 5.28 & 0.37 & 48.17\tabularnewline
 & Ours & 7.41 & 0.39 & 48.20\tabularnewline
\midrule[0.10em] 
\multirow{2}{*}{Res-50} & Baseline & 6.43 & 0.40 & 48.84\tabularnewline
 & Ours & 8.04 & 0.41 & 48.86\tabularnewline
\midrule[0.10em] 
\multirow{2}{*}{Res-101} & Baseline & 8.24 & 0.43 & 50.31\tabularnewline
 & Ours & 10.17 & 0.45 & 50.37\tabularnewline
\bottomrule[0.15em] 
\end{tabular}}
\vspace{-5mm}
\end{table}

\section{More qualitative results}
\vspace{0mm}

\begin{figure*}[t]
    \centering{}
     \includegraphics[scale=0.17]{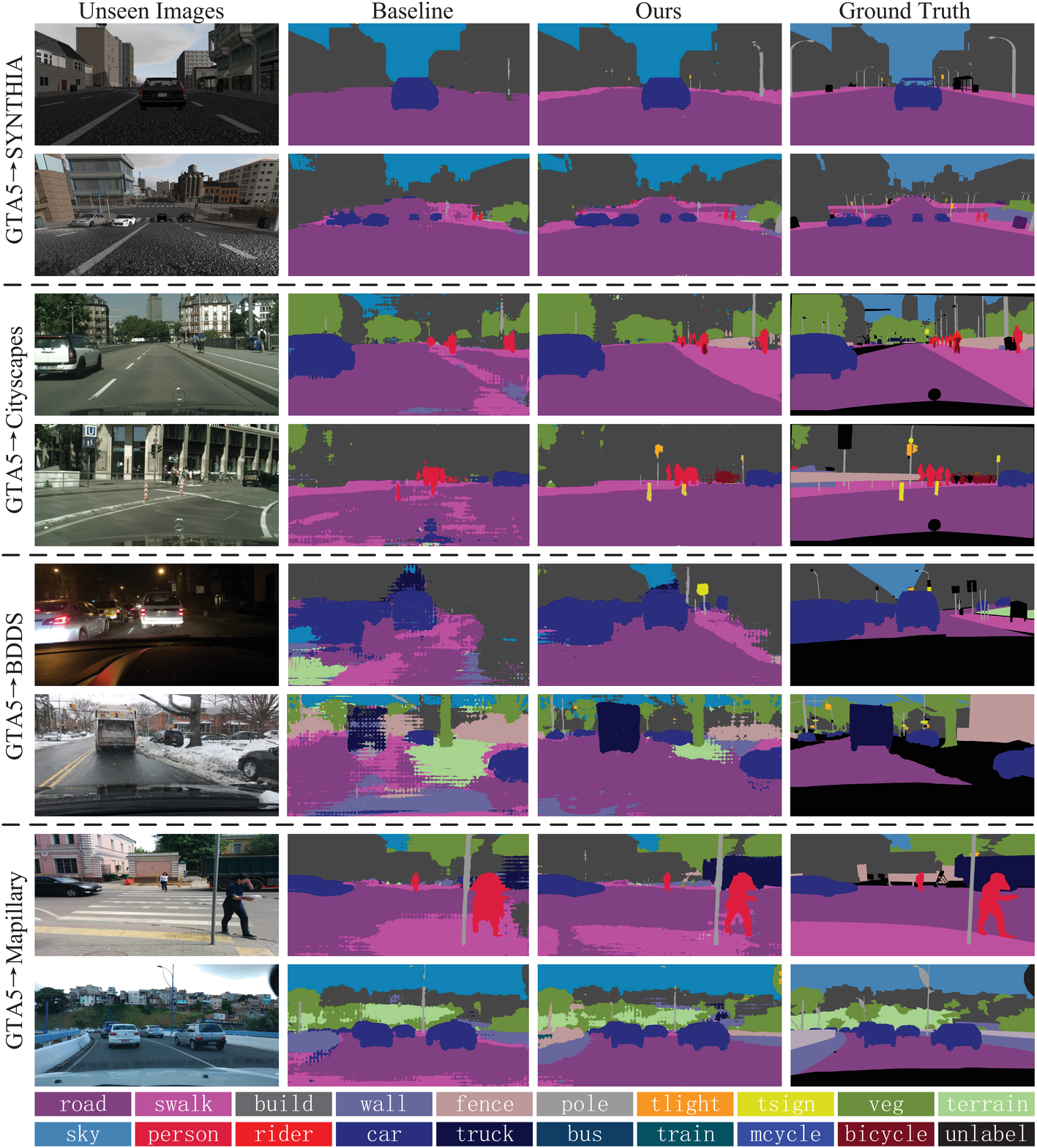} 
     \vspace{-3mm}
     \caption{Qualitative results of our approach generalizing from GTA5 to other four domains. \label{fig: more qualitative results}
     }
     \vspace{-3mm}

    \end{figure*}

\cref{fig: more qualitative results} shows more qualitative results under various unseen domains.  We demonstrate the effects of the proposed semantic-aware feature matching by comparing the segmentation results from our proposed approach and the baseline.
In the setting of GTA5 $\rightarrow$ Mapillary, the baseline fails to cope with these weather changes, while ours still shows fair results. Under the illumination changes as shown in GTA5 $\rightarrow$ BDDS, our method finds the road and sidewalk clearer than the baseline.

%--------------------------figure 2
%--------------------------figure 2

\balance
%%%%%%%%% REFERENCES
{\small
\bibliographystyle{ieee_fullname}
\bibliography{egbib}
}